\documentclass[letterpaper]{article}

\usepackage[final]{eurips_2025}

\usepackage{graphicx}
\usepackage{amsmath, amssymb, amsthm, bm}
\usepackage{mathtools}
\usepackage{booktabs}
\usepackage{bbm}
\usepackage[font=small,labelfont=bf]{caption}
\usepackage{subcaption}

\usepackage{enumitem}
\newcommand{\E}{\mathbb{E}}
\newcommand{\Var}{\mathbb{V}}
\newcommand{\Cov}{\mathbb{C}\mathrm{ov}}

\newcommand{\x}{\bm{x}}

\newcommand{\w}{\bm{w}}

\newcommand{\thetaV}{\bm{\theta}}

\newcommand{\MSE}{\mathrm{MSE}}
\newcommand{\EV}{\mathrm{EV}}
\newcommand{\Ggap}{G_{\Delta}}                % structural gap (bias-squared term)
\newcommand{\Rec}{\mathrm{rec}}
\newcommand{\Dir}{\mathrm{dir}}

\newcommand{\Tmap}{\mathcal{T}}                % composition map from 1-step params to h-step
\newcommand{\Jh}{J_h}                % Jacobian of T at h
\newcommand{\Th}{T_h}                % scalar amplification summary

\theoremstyle{definition}

\theoremstyle{plain}
\newtheorem{proposition}{Proposition}
\newtheorem{theorem}{Theorem}
\newtheorem{lemma}{Lemma}
\theoremstyle{remark}

\title{Epistemic Error Decomposition for Multi-step Time Series Forecasting: 
Rethinking Bias–Variance in Recursive and Direct Strategies}

\author{
  Riku Green$^{1}$\thanks{Correspondence: riku.green@bristol.ac.uk} \quad
  Huw Day$^{1}$ \quad
  Zahraa S.~Abdallah$^{1}$ \quad
  Telmo M.~Silva Filho$^{1}$ \\
  \\
  $^{1}$University of Bristol
}

\begin{document}
\maketitle

\begin{abstract}
Conventional wisdom in multi-step forecasting claims that recursive strategies are ``high bias, low variance'' while direct strategies are ``low bias, high variance''. We revisit this claim through a decomposition of expected multi-step forecast error into (i) aleatoric/irreducible noise, (ii) a epistemic structural gap $\Ggap$ (approximation bias), and (iii) an epistemic estimation-variance term $\EV$. Our analysis shows that: (1) with \emph{linear predictors}, independent of the data, $\Ggap = 0$; however (2) for \emph{nonlinear predictors} the composition inherent to recursion can increase expressivity, making $\Ggap$ sign-indefinite, and data dependent; and (3) the multistep estimation variance of the recursive strategy ($\EV_{\Rec}$) factorizes as the one-step $\EV_1$ multiplied by a Jacobian-derived amplification $\Th$ of the parameter-composition map. This decomposition highlights conditions where recursive strategies can have both lower bias and also higher variance than direct strategies. We reproduce the same phenomenon empirically for the MLPs on the commonly known ETTm1 dataset. This perspective provides actionable guidance for selecting forecasting strategies based on model nonlinearity and noise characteristics, rather than relying on heuristic bias–variance assumptions.
\end{abstract}

% =======================
% 1. Introduction
% =======================
\section{Introduction}
Multi-step time-series forecasting (TSF) underpins applications such as energy demand, traffic, and finance, where future values must be predicted several steps ahead \citep{lim2021time}. 
For a forecast horizon $h$, two dominant strategies exist: (a) \emph{recursive}, training a one-step model and iterating it $h$ times, and (b) \emph{direct}, training separate $h$-ahead predictors.
A prevailing narrative states that recursion suffers more bias but less variance, while direct methods exhibit the opposite trade-off \citep{taieb2014machine,Green2025Stratify}. 
In practice, however, recursive strategies can perform on par with direct ones, or even significantly better \citep{Green2025Stratify}, despite direct strategies appearing theoretically preferable. These findings motivate further research in better understanding the underlying factors that drive this behaviour. We argue that a precise error decomposition involving the Jacobian of the recursive strategy's composition provides insight into these unexplained empirical findings. This work represents a step toward clarifying how epistemic uncertainty arises in multi-step forecasting, offering guidance for model design and selection.

\paragraph{Contributions.}
\begin{enumerate}[leftmargin=*]
  \item \textbf{Nonlinear recursion alters expressivity and can invert bias ordering.} 
  % By expanding the composition, for non-linear classes, we show that recursive strategies are not always higher bias.
  Recursive composition expands the representable function space, making the bias difference between recursive and direct predictors data-dependent rather than fixed.

  \item \textbf{A Jacobian-based decomposition of estimation variance.} 
  % Using a delta method, we show that the finite sample variance of the recursive strategy depends heavily on the Jacobian norm of the composition map resulting from autoregressive forecasting. This result theoretically explains where recursive strategies can gain higher estimation variance than direct strategies.
  % Using a delta-method expansion of the parameter composition map, we derive a closed-form relation linking recursive and one-step estimation variances, identifying the Jacobian norm as the amplification factor for epistemic uncertainty.
  Using the delta method, we show that recursive variance equals one-step variance scaled by a Jacobian-derived amplification factor, linking parameter geometry to uncertainty growth.

  \item \textbf{Characterisation of noise regimes.} 
  % The interaction between aleatoric noise terms (measurement noise and process noise) and the data generating process (DGP) have a strong relationship with the expected variance of MSF strategies. We show that cases where process noise is relatively higher than measurement noise favour the recursive strategy and the converse for direct strategies.
  By separating process and measurement noise, we show that their relative magnitudes govern the expected estimation variance of each strategy, yielding a practical diagnostic for strategy selection.

\end{enumerate}

\section{Background}
\label{background}

We consider a stationary latent process $\{x_t\}$ with observations $y_t = x_t + v_t$ (measurement noise). 
A recursive strategy trains a one-step predictor $\hat{f}_1$ mapping recent observations to the next step, for example
$\hat{y}_{t+1} = \hat{f}_1(y_t, y_{t-1}, \ldots, y_{t-p+1})$ using $p$ lagged inputs. 
The $h$-step predictor $\hat{f}_h = \hat{f}_1^{\circ h}$ is then obtained by repeatedly applying $\hat{f}_1$, 
feeding each intermediate prediction back as an input. 
A direct strategy instead trains a separate model $\hat{g}_h$ to predict $y_{t+h}$ directly from the same $p$ lagged features used by $\hat{f}_1$. 
Their performance can be compared through the mean squared error (MSE).

We distinguish three predictors:
(i) the Bayes-optimal $f_h^\star$ that minimizes the true expected risk;
(ii) the best-in-class oracle $f_h^{\mathcal H}$ within a hypothesis space 
$\mathcal H\in\{\mathrm{Rec},\mathrm{Dir}\}$;
and (iii) the finite-sample estimator $\hat f_h^{\mathcal H}$ obtained from $N$ training samples.
The expected test MSE of $\hat f_h^{\mathcal H}$ decomposes as
\begin{equation}
  \E[\MSE(\hat f_h^{\mathcal H})]
  = \underbrace{\sigma^2_{\text{noise}}}_{\text{irreducible noise}}
  + \underbrace{G_{\mathcal H}}_{\text{structural gap / bias}}
  + \underbrace{EV_{\mathcal H}}_{\text{estimation variance}},
  \label{eq:mse-decomp}
\end{equation}
where 
$\sigma^2_{\text{noise}}=\E[(y_{t+h}-f_h^\star)^2]$ is the irreducible aleatoric uncertainty,
$G_{\mathcal H}=\E[(f_h^\star-f_h^{\mathcal H})^2]$ 
is the approximation bias of class~$\mathcal H$, 
and 
$EV_{\mathcal H}=\E[(f_h^{\mathcal H}-\hat f_h^{\mathcal H})^2]$
is the finite-sample estimation variance.

The class-level (oracle) MSE is 
$\sigma_{\epsilon,h}^2=\E[(y_{t+h}-f_h^{\mathcal H})^2]
 =\sigma^2_{\text{noise}}+G_{\mathcal H}$,
which differs from the purely irreducible term $\sigma^2_{\text{noise}}$.
To compare strategies we analyse
$\Ggap = G_{\mathrm{Rec}}-G_{\mathrm{Dir}}$ 
and 
$EV_{\Delta}= EV_{\mathrm{Rec}}-EV_{\mathrm{Dir}}$.

% \section{Related Work}
% \label{sec:related}
\paragraph{Related Work}

% \textbf{Multi-step strategies.} 
Multi-step time series forecasting (MSF) strategies, such as recursive and direct significantly impact performance, with no universally optimal approach \cite{green2024time, Green2025Stratify, taieb2014machine, petropoulos2022forecasting}. Recursive methods accumulate prediction errors across steps, whereas direct methods avoid this accumulation by training independent models for each forecast horizon. Recent foundation model designs for TSF also have not resolved which strategy to adopt \cite{miller2024survey}.
\textbf{Error decompositions.}
Bias–variance decompositions are widely used to characterise epistemic uncertainty and trace model errors \cite{hullermeier2021aleatoric, zhou2022survey}. 
However, prior analyses do not analyse the Jacobian of composition functions in expressing recursive strategies as direct equivalents \cite{taieb2014machine, petropoulos2022forecasting}.
\textbf{Noise modeling.} 
While process and measurement noise are central to real-world settings, such as transport network forecasting \cite{rodrigues2018heteroscedastic}, and in Kalman filters \cite{bilgin2025joint}, their its role in MSF strategy selection remains underexplored. We address this in Section \ref{se:noise}.
\textbf{Research gap.} 
In summary, we decompose MSF epistemic uncertainty via bias-variance terms and leveraging Jacobians for variance amplification (Section \ref{sec:bias} and \ref{sec:ev}) and consider these practical noise regimes. This provides theoretical insights into machine learning behaviour in TSF and towards actionable guidance for strategy selection.

% \newpage

% =======================
% 4. Bias / Structural Gap G
% =======================
\section{Structural Gap $\Ggap$: When is Recursive More Biased?}
\label{sec:bias}

We begin by analysing the linear predictor class, recovering a known result where recursion and direct forecasting are equivalent in structural bias \citep{taieb2014machine}. We then extend the argument to nonlinear predictors, where the gap becomes data-dependent.

\subsection{Linear predictor classes: $\Ggap = 0$}

\begin{figure}[t!]
    \centering
    \begin{minipage}[b]{0.29\textwidth}
        \subcaptionbox{%
        }[1\linewidth]{%
        \includegraphics[width=\textwidth]{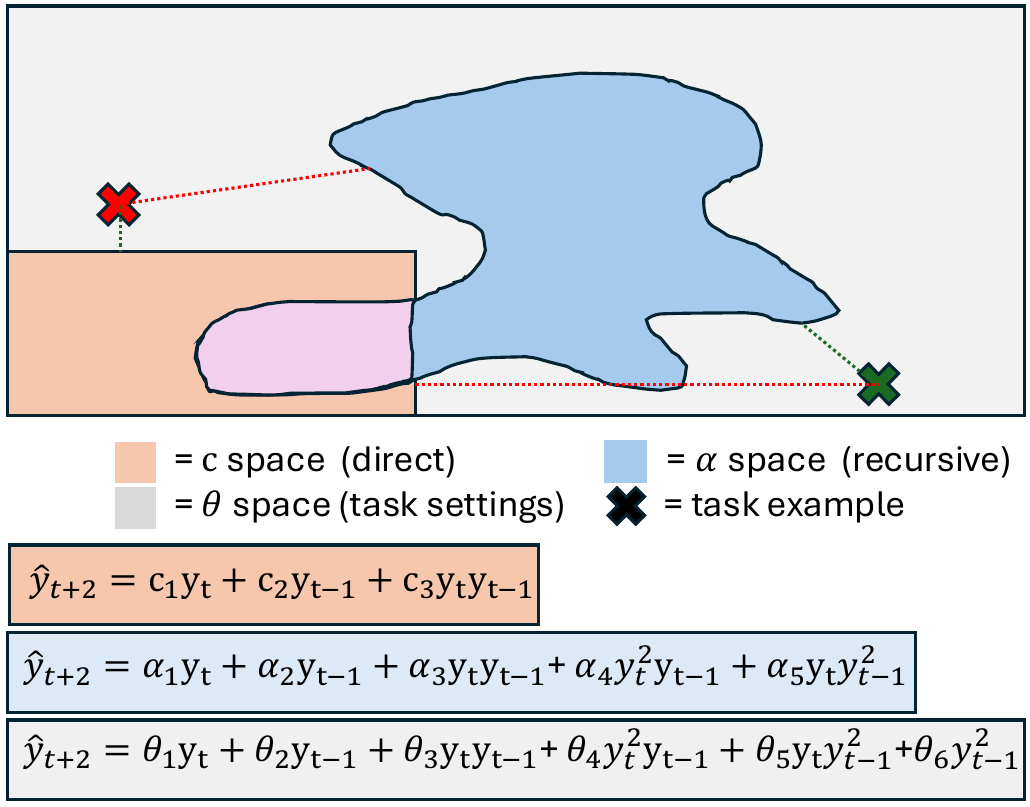}}
    \end{minipage}
    \hfill
    \begin{minipage}[b]{0.32\textwidth}
        \subcaptionbox{%
        }[1\linewidth]{%
        \includegraphics[width=\textwidth]{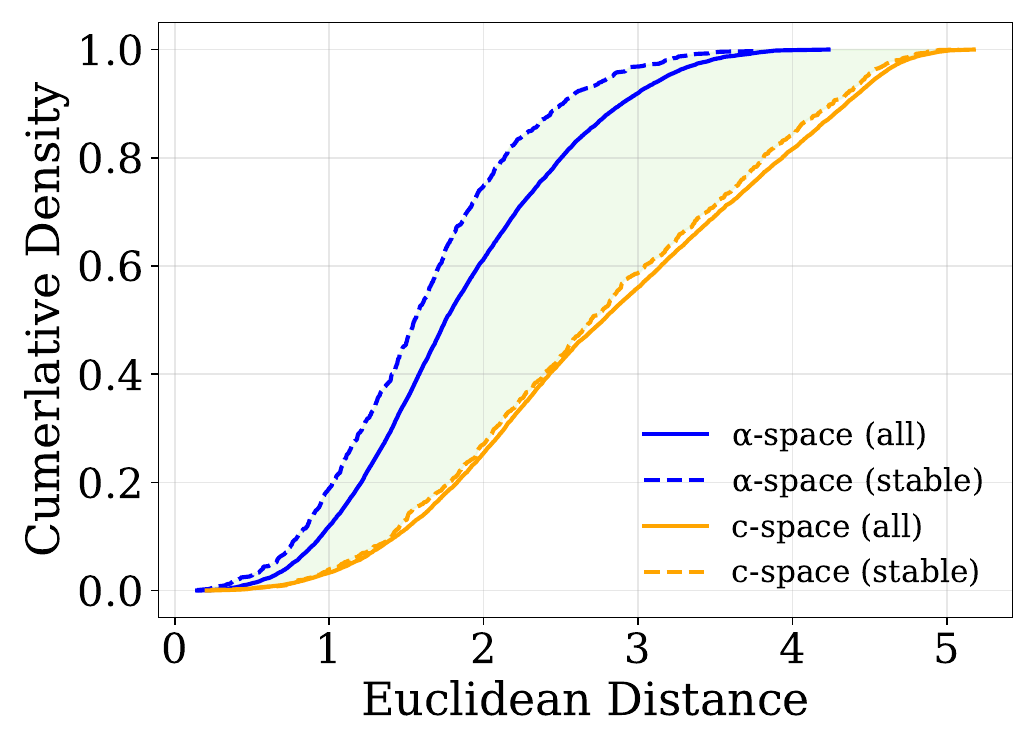}}
    \end{minipage}
    \hfill
    \begin{minipage}[b]{0.32\textwidth}
        \subcaptionbox{%
        }[1\linewidth]{%
        \includegraphics[width=\textwidth]{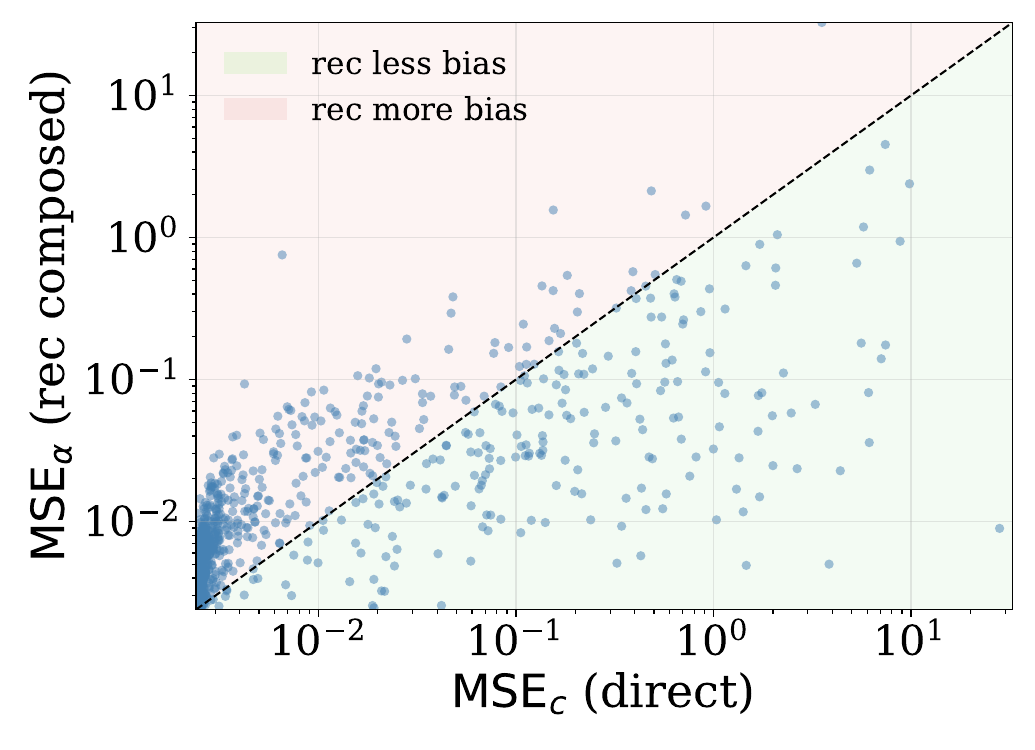}}
    \end{minipage}
    \caption{%
    \textbf{Recursive composition expands the representable function space and can reduce bias.}
    (a)~Schematic: direct strategies (in $c$) span a 3D subspace, while recursive strategies (in $\alpha$ under recursive composition) occupy a richer 5D subspace.  We show a 6D task manifold (in \(\theta\)). Green or red task show $\alpha$ is closer or further to task setting, respectively.
    (b)~Empirical coefficient–structure bias: recursive models (\(\alpha\)-space) lie systematically closer to uniformly sampled task parameters than direct ones (\(c\)-space).
    (c)~Pairwise MSE comparison: recursive predictors can—but do not necessarily—achieve lower bias than direct models.
    }
    \label{fig:bilinear_G_comp}
\end{figure}

\paragraph{Recursive mapping from $(b_1,b_2)$ to $(\alpha_1,\alpha_2)$.}
Consider a one–step linear predictor defined by:
\begin{equation}
    \hat{y}_{t+1} = b_1 y_t + b_2 y_{t-1}.
\end{equation}
By composing this relation one additional step, we obtain a two–step prediction:
\begin{align}
    \hat{y}_{t+2}
        &= b_1 y_{t+1} + b_2 y_t
         = b_1(b_1 y_t + b_2 y_{t-1}) + b_2 y_t 
        = (b_1^2 + b_2)\, y_t + (b_1 b_2)\, y_{t-1}.
\end{align}

Hence, the effective two–step coefficients can be written as 
\([\alpha_1,\, \alpha_2]^{\mathsf{T}} = [\,b_1^2 + b_2,\, b_1 b_2\,]^{\mathsf{T}} = A(b_1, b_2),\)
where the nonlinear map \(A : \mathbb{R}^2 \!\to\! \mathbb{R}^2\) encodes the
composition of the one–step predictor with itself.

\paragraph{Result (surjectivity of linear predictor compositions).}
Consider the one–step linear model $\hat y_{t+1}=b_1 y_t+b_2 y_{t-1}$ and its $h{=}2$ recursive composition
$A:\mathbb{R}^2\!\to\!\mathbb{R}^2,\; A(b_1,b_2)=(\alpha_1,\alpha_2)$ with
$\alpha_1=b_1^2+b_2,\ \alpha_2=b_1 b_2$.
\textbf{Then $A$ is surjective}: for every $(\alpha_1,\alpha_2)\in\mathbb{R}^2$ there exists at least one $(b_1,b_2)\in\mathbb{R}^2$ such that $A(b_1,b_2)=(\alpha_1,\alpha_2)$.
(\emph{Proof deferred to Appendix~\ref{surjective_proof}}.)

Surjectivity implies that there always exists a one-step linear model that is equivalent to the direct two-step linear model;
hence the structural gap vanishes: $G_{\mathrm{rec}} \;=\; G_{\mathrm{dir}} \implies \Ggap = 0$.

% Therefore any performance differences between recursive and direct \emph{linear} AR(2) predictors at $h{=}2$ arise from estimation effects (variance/regularization), not representational bias.

\subsection{Nonlinear predictor classes: $\Ggap \in \Re$}
For nonlinear one-step predictors, the recursive composition map is no longer surjective onto the space of direct predictors. We illustrate this with a simple example. Consider the one–step model:
\begin{equation}
    \widehat{y}_{t+1}
    = b_1 y_t + b_2 y_{t-1} + b_3 y_t y_{t-1},
\end{equation}
where $b_3$ controls the strength of a simple bilinear interaction.
By composing this map once, we can express the two–step prediction entirely
in terms of $(y_t, y_{t-1})$:
\begin{equation}
    \widehat{y}^{\mathrm{Rec}}_{t+2}
    = \alpha_1 y_t + \alpha_2 y_{t-1}
      + \alpha_3 y_t y_{t-1}
      + \alpha_4 y_t^2
      + \alpha_5 y_t^2 y_{t-1},
\end{equation}
% where the coefficients follow the nonlinear transformation
\begin{align*}
    \text{where }
    \alpha_1 &= b_1 + b_2^2, &
    \alpha_2 &= 2 b_1 b_2, &
    \alpha_3 &= b_3 (b_1 + b_2), &
    \alpha_4 &= b_3 b_1, &
    \alpha_5 &= b_3^2.
\end{align*}
Thus the recursive composition induces a nonlinear mapping
$A : (b_1,b_2,b_3) \mapsto (\alpha_1,\ldots,\alpha_5)$.

In contrast, a direct two–step model of the same polynomial order would learn
\begin{equation}
    \widehat{y}^{\mathrm{Dir}}_{t+2}
    = c_1 y_t + c_2 y_{t-1} + c_3 y_t y_{t-1},
\end{equation}
whose parameterization $(c_1,c_2,c_3)$ spans three independent dimensions of a
subset of the corresponding feature space, whereas the recursive model operates over a constrained submanifold of a larger five dimensional subspace induced by the composition map
$A$. \\
\textbf{Empirical validation.}
Figure \ref{fig:bilinear_G_comp} compares the distance in coefficient space between each strategy’s best-in-class predictor and the ground-truth function, showing that the relative bias of recursive and direct models varies with the data. Experimental details are outlined in Appendix ~\ref{fig_1_details}.

\paragraph{Structural implication.}
In the linear case, we showed $\Ggap = 0$. 
For nonlinear predictors, this equality no longer holds because the composition changes the effective function space.
The structural gap $\Ggap$ therefore reflects differences in expressivity between the recursive and direct classes: depending on the data-generating process, nonlinear interactions can either constrain or enrich the span of the recursive model, making $G_{rec} - G_{dir}$ data-dependent in sign.

\begin{figure}
    \centering
    \includegraphics[width=\linewidth]{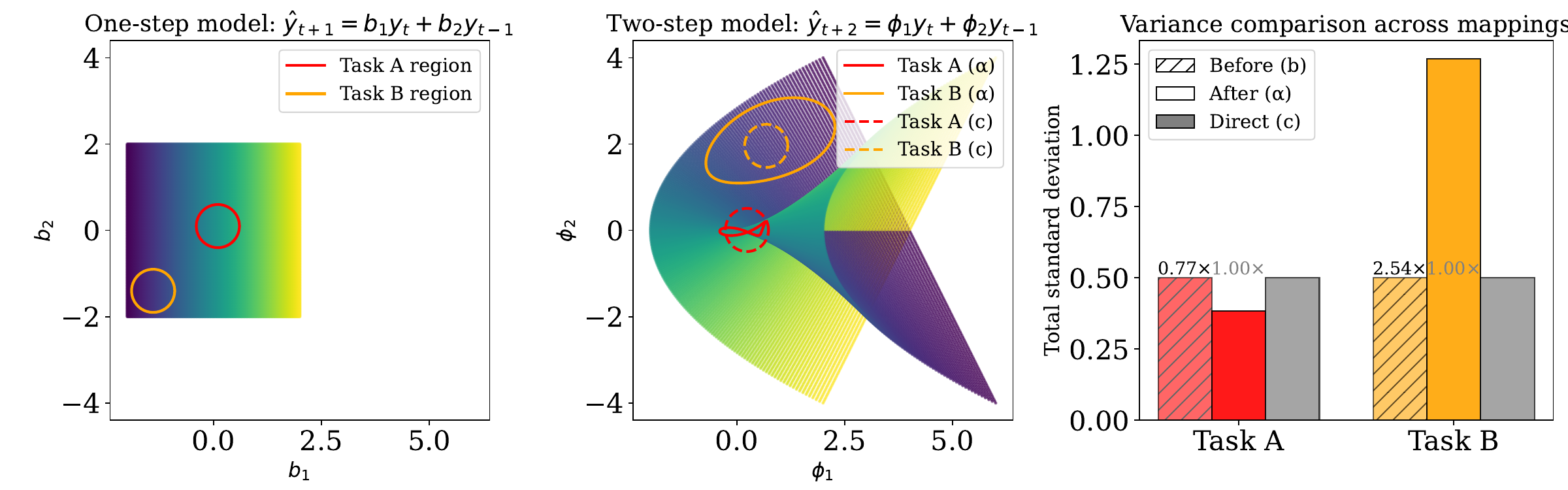}
\caption{
\textbf{Mapping geometry and variance propagation between one-step and two-step forecasting models.}
\textit{Left:} one-step linear parameter space $(b_1,b_2)$ for two synthetic tasks (A, B), each defining a local region of fitted one-step models $\hat{y}_{t+1}=b_1y_t+b_2y_{t-1}$.
\textit{Middle:} corresponding two-step coefficient space $(\phi_1,\phi_2)$ obtained via recursive composition (solid ellipses, $\alpha$) and via direct two-step fitting (dashed ellipses, $c$). 
The nonlinear mapping from $b$-space to $\phi$-space distorts local geometry, leading to either amplification or contraction of parameter variance depending on position.
\textit{Right:} total coefficient variance before mapping ($b$), after recursive mapping ($\alpha$), and for directly fitted models ($c$). 
For illustration, we assume equal variance in the direct coefficients to the one-step task which is likely to vary in practice.
}
    \label{fig:geometry map}
\end{figure}
% \newpage

% =======================
% 5. Estimation Variance (EV) via Jacobians
% =======================
% core idea: at least linear relationship between coef variance and mse variance.
% coef variance at 1 step < at h step
% if the composition stretches the 1 step coef variance 
% then the effective coef variance can be > h step
% which would guarentee higher mse variance by monotonicity

\section{Estimation Variance $\EV$: Jacobian Amplification for Recursion}
\label{sec:ev}
We view recursion as a composition map and apply the delta method to quantify how estimation uncertainty propagates across steps. This yields a Jacobian-based expression for epistemic variance, which we later relate to aleatoric noise.

\subsection{A composition map from one-step to $h$-step}
Let $\thetaV^\star$ be the population-optimal one-step parameters in $\mathcal{F}_1$, and let $\hat{\thetaV}$ be a finite-sample estimator with covariance $\Sigma_\theta = \Cov(\hat{\thetaV})$. 
Assume the $h$-fold recursion $f_1^{\circ h}$ 
\[
  \Tmap_h:\ \thetaV \mapsto \alpha_h(\thetaV),
\]
which parameterizes the \emph{effective} $h$-ahead model. 
If the one-step predictor $\hat{f}_{1}$ is smooth and differentiable everywhere,
then under chain rule it follows that its $h$-fold composition $\hat{f}_h$ is also continuously differentiable. This allows us to apply a first-order Taylor expansion approximation.

Applying a first-order Taylor expansion (delta method) about $\thetaV^\star$,
\[
  \alpha_h(\hat{\thetaV}) 
  \;=\; \alpha_h(\thetaV^\star) 
  + \Jh(\thetaV^\star)\,(\hat{\thetaV}-\thetaV^\star) 
  + R_2,
\]

where $\Jh(\thetaV^\star)$ is the Jacobian matrix and $R_2$ collects higher-order terms in $(\hat{\thetaV}-\thetaV^\star)$, proof shown in Appendix~\ref{jacob_proof}. 
When local curvature is modest and estimation noise is small so that $R_2$ is negligible, the induced parameter covariance at horizon $h$ satisfies:
\begin{equation}
  \label{eq:alpha-cov}
  \Sigma_{\alpha_h} \;\approx\; \Jh\, \Sigma_\theta\, \Jh^\top.
\end{equation}
Figure \ref{fig:geometry map} shows that, even for a linear predictor, the Jacobian of its composition map directly scales the estimation variance over coefficients of the one-step model into the resulting composed model.

\begin{figure}[t!]
    \centering
    \begin{minipage}[b]{0.32\textwidth}
        \centering
        \includegraphics[width=\textwidth]{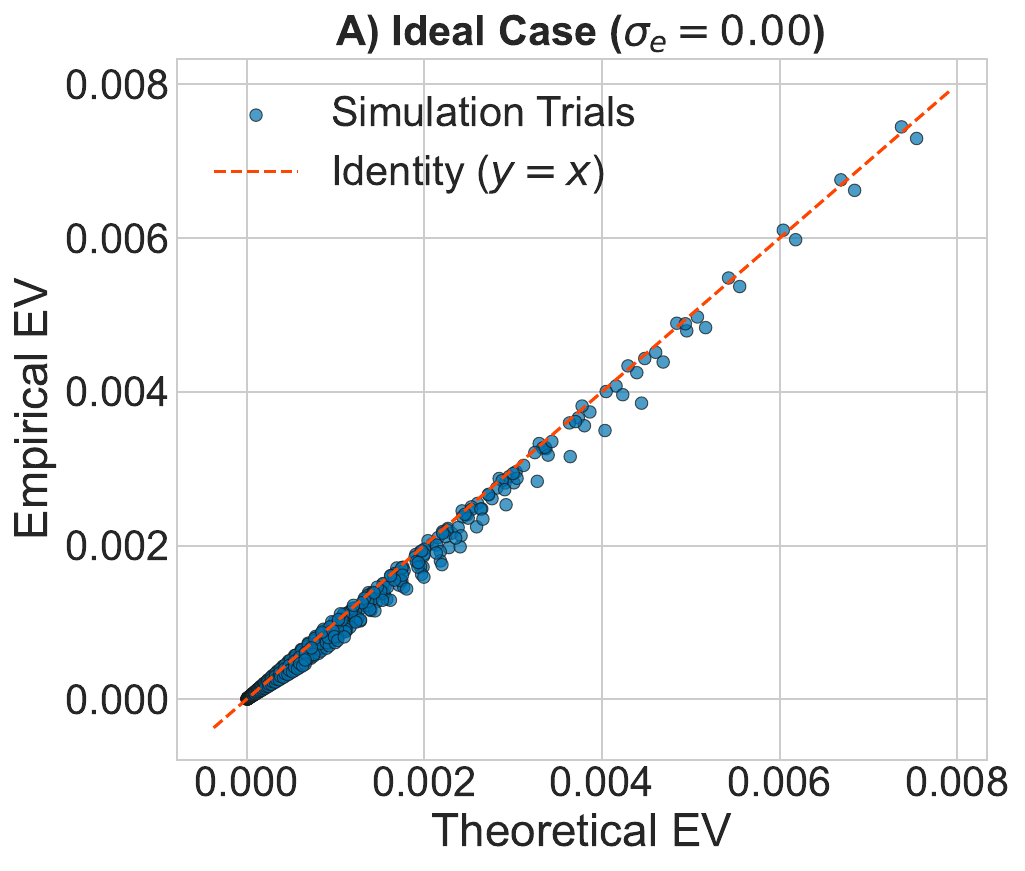}
    \end{minipage}
    \hfill
    \begin{minipage}[b]{0.32\textwidth}
        \centering
        \includegraphics[width=\textwidth]{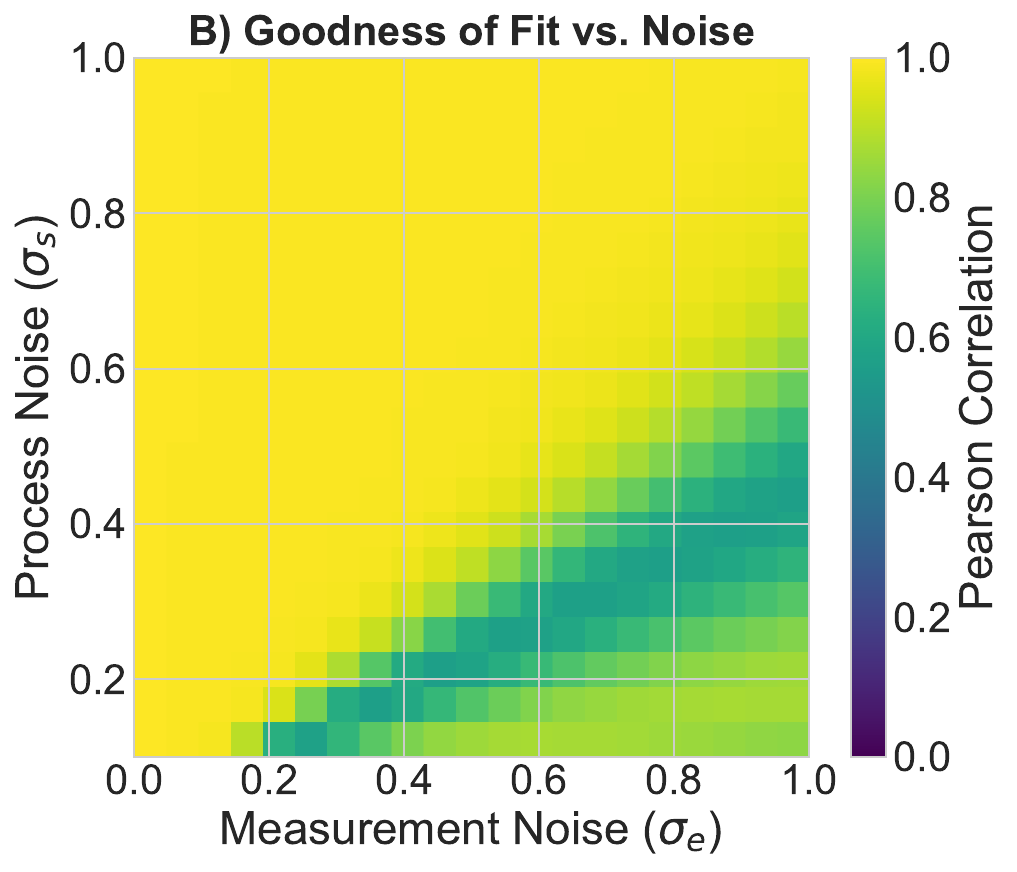}
    \end{minipage}
    \hfill
    \begin{minipage}[b]{0.32\textwidth}
        \centering
        \includegraphics[width=\textwidth]{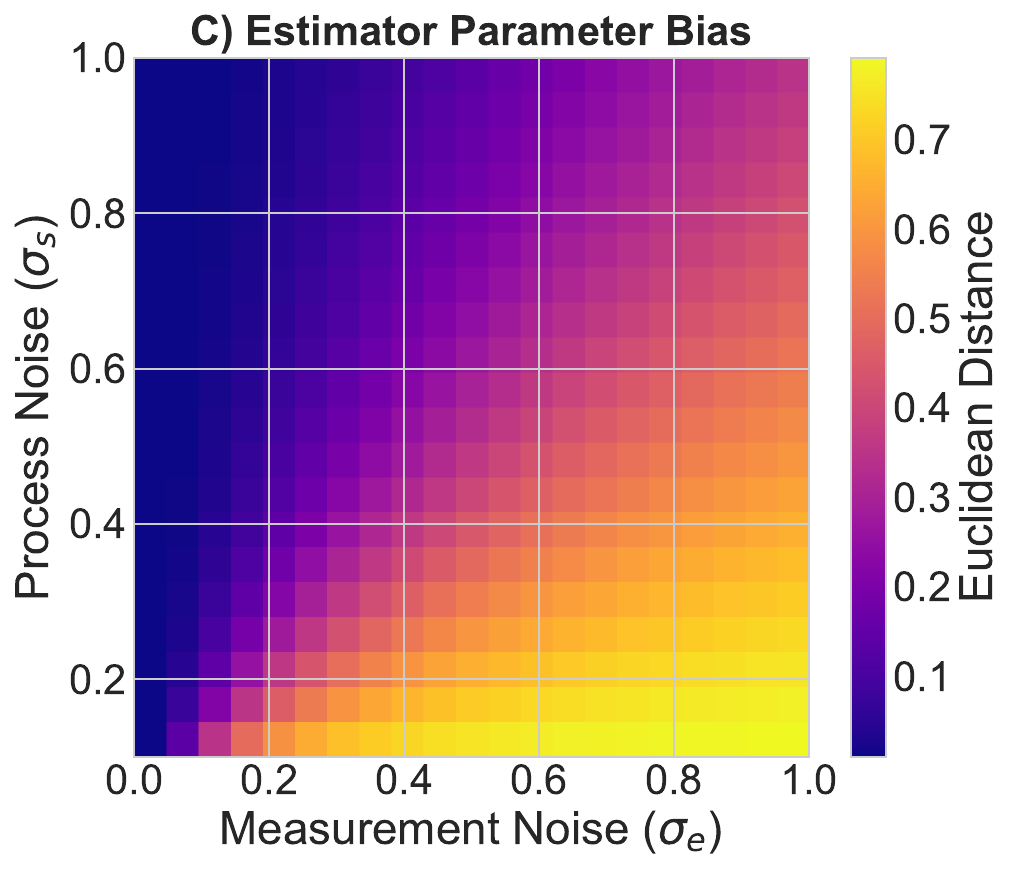}
    \end{minipage}

\caption{
        \textbf{Empirical Validation of Theoretical EV and the Impact of Estimator Bias.}
        This figure analyzes our theoretical model's performance by sweeping across stable AR(2) parameters ($a, \gamma$) and varying process ($\sigma_s$) and measurement ($\sigma_e$) noise.
        \textbf{(A)} In the ideal case with minimal measurement noise ($\sigma_e \approx 0$), the OLS estimator is unbiased. Here, the empirical EV aligns almost perfectly with our theoretical prediction, validating the fundamental accuracy of the Jacobian-based formulation.
        \textbf{(B)} This panel maps the theory's goodness-of-fit by plotting the Pearson correlation between theoretical and empirical EV for each noise configuration. A clear degradation is visible as measurement noise domninates process noise.
        \textbf{(C)} This panel diagnoses the root cause of the breakdown. It visualizes the parameter bias of the OLS estimator, measured as the Euclidean distance between the true DGP parameters and the average converged coefficients. The heatmap shows that the estimator bias grows significantly with measurement noise, mirroring the pattern of degradation seen in panel (B).
    }

    \label{fig:sigma_e}
\end{figure}

\subsection{EV comparison: recursive vs direct}

As shown in Eq.~\eqref{eq:alpha-cov}, parameter uncertainty at horizon $h$ is propagated by the composition Jacobian $\Jh$, yielding $\Sigma_{\alpha_h} \approx \Jh\,\Sigma_\theta\,\Jh^\top$. 
To translate this parameter covariance into test MSE variance, we separate the \emph{base} feature geometry from the \emph{composition}-induced geometry. 
Let $\x=[y_t, y_{t-1},\dots, y_{t-p+1}]$ denote the one–step (and direct) features with second moment
\[
  Q \;:=\; \E[\x \x^\top],
\]
which is \emph{shared} by one–step and direct models. 
Under recursion, composing the one–step predictor $h$ times induces effective $h$–ahead features $\tilde{\x}$ with second moment
\[
  \widetilde{Q} \;:=\; \E[\tilde{\x}\,\tilde{\x}^\top],
\]
which generally differs from $Q$ because composition generates new feature interactions. 
Intuitively, $\Sigma_{\alpha_h}$ quantifies “how uncertain the effective parameters are,” while $\widetilde{Q}$ (or $Q$ for direct) ``encodes how data geometry maps that uncertainty into prediction space".

For linear DGPs and well-specified linear-in-parameters predictors, these second moments yield exact EV expressions; for nonlinear DGPs/predictors, the equalities below are first-order (delta-method) approximations whose quality depends on higher-order terms.

\begin{theorem}[Jacobian-driven EV for recursion]
\label{thm:jacobian-ev}
Under standard regularity for linear-in-parameters models with exogenous regressors, the recursive $h$-ahead estimation-variance satisfies
\[
  \boxed{\;
  \EV_{\Rec}(h) \;\approx\; 
  \mathrm{tr}\!\big(\Jh\, \Sigma_\theta\, \Jh^\top \,\widetilde{Q}\big).
  \;}
\]
Let the one–step baseline be $\EV_{1\text{-step}}=\mathrm{tr}(\Sigma_\theta Q)$. 
Define the (dimensionless) amplification
\[
  T_h \;\equiv\; 
  \frac{\mathrm{tr}\!\big( \Jh\, \Sigma_\theta\, \Jh^\top \,\widetilde{Q} \big)}
       {\mathrm{tr}\!\big(\Sigma_\theta\, Q\big)}
  \qquad\Rightarrow\qquad
  \EV_{\Rec}(h)\;\approx\; T_h\cdot \EV_{1\text{-step}}.
\]
\end{theorem}

\noindent\textbf{Corollary (EV difference).}
If the direct $h$–ahead estimator has covariance $\Sigma_\theta^{(h)}$ on the \emph{same} base features (both $\hat{g}_h, \hat{f}_1:\x\rightarrow \Re$ as from Section \ref{background}, so its geometry is $Q$), then
\[
  \Delta \EV(h) 
  \;:=\; \EV_{\Rec}(h) - \EV_{\Dir}(h)
  \;\approx\; 
  \mathrm{tr}\!\big(\Jh\,\Sigma_\theta\,\Jh^\top \,\widetilde{Q}\big)
  \;-\;
  \mathrm{tr}\!\big(\Sigma_\theta^{(h)}\, Q\big).
\]

\begin{proof}[Proof sketch]
See Appendix~\ref{ev_proof}. \textit{Note on nonlinear predictors also in Appendix.}
\end{proof}

% \textbf{Noise enters in two places.}
% Noise influences our decomposition through two distinct pathways, with measurement noise playing a critical role in the \emph{perceived} bias of recursive strategies. While process noise ($\sigma_s^2$) perturbs the latent dynamics, measurement noise ($\sigma_e^2$) fundamentally alters the estimation problem. Figure~\ref{fig:sigma_e} empirically supports our theory for a linear DGP and linear predictor, and also highlights a failure mode: measurement noise misspecifies the predictive model, preventing $\E[\hat{\theta}] \to \theta^\star$, necessary for the linearisation accuracy. We can show that, for a linear DGP, the aleatoric noise terms admit the closed forms
% \[ \begin{aligned} \quad & \sigma_{noise, 1}^2 = \sigma_s^2 + (1+a^2+\gamma^2)\sigma_e^2, &\qquad \quad & \sigma_{noise, 2}^2 = (1+a^2)\sigma_s^2 + \big[1+(a^2+\gamma)^2+(a\gamma)^2\big]\sigma_e^2. \end{aligned} \] (\emph{Proof deferred to Appendix~\ref{noise_proof}}.)
% Hence, depending on $(a,\gamma,\sigma_s,\sigma_e)$, the aleatoric noise of the optimal one-step task can exceed that of the two-step task.
% Empirically, larger $\sigma_s/\sigma_e$ tends to favour recursion, and the converse favouring the direct approach.

\textbf{Noise enters in two places.}
\label{se:noise}
Figure~\ref{fig:sigma_e} illustrates how process and measurement noise affect both the \emph{aleatoric floor} and the \emph{epistemic estimation variance}. 

% \textbf{(A)} When measurement noise is negligible, the empirical estimation variance aligns almost perfectly with the Jacobian-based theoretical prediction, validating the accuracy of our formulation in the well-specified regime.  
% \textbf{(B)} As measurement noise increases relative to process noise, this agreement degrades: the Pearson correlation between theoretical and empirical EV decreases, revealing that measurement noise disrupts the model’s assumptions.  
% \textbf{(C)} The third panel diagnoses this breakdown by visualizing estimator bias, measured as the Euclidean distance between the true DGP parameters and the mean estimated coefficients. The bias grows rapidly with $\sigma_e^2$, mirroring the degradation observed in (B). This demonstrates that measurement noise misspecifies the regression problem, preventing $\E[\hat{\theta}] \to \theta^\star$ and causing the theoretical EV to underestimate the true uncertainty.
We can show that, for a linear DGP, the aleatoric noise terms admit the closed forms
\[ \begin{aligned} \quad & \sigma_{noise, 1}^2 = \sigma_s^2 + (1+a^2+\gamma^2)\sigma_e^2, &\qquad \quad & \sigma_{noise, 2}^2 = (1+a^2)\sigma_s^2 + \big[1+(a^2+\gamma)^2+(a\gamma)^2\big]\sigma_e^2. \end{aligned} \] (\emph{Proof deferred to Appendix~\ref{noise_proof}}.)
These expressions show that, depending on $(a,\gamma,\sigma_s,\sigma_e)$, even the intrinsic one-step task can exhibit greater aleatoric uncertainty than the two-step task. Empirically, larger $\sigma_s / \sigma_e$ ratios tend to favour recursion, whereas dominant measurement noise favours the direct approach (see Appendix \ref{fig:sigma_e} and its Figure \ref{fig:analytic_empirical_comparison}).

% % noise doesnt affect the jacobian compuation
% % noise affects Sigma which affects the size of T
% % noise affects sigma_epsilons.
% % can plot the ratio of parity over sigma_e,s values
% % sigma_eps_h scales like sigma_e + scalar * sigma_s

% =======================
% 6. Empirical Study
% =======================
\section{Empirical Study: Replication on real-world data and MLP predictors}
\label{sec:empirical}
% what we see is that the ratio of sigma_eps_1/h is dependent on the noise type ratio
% We do not know the noise types
% generally, the ev ratio is below zero
% so we can focus on estimating T
% our T estimate gets better with sample size
% if we set a decision rule of T<2 try recursive, we can see how often this wins when we sweep over DGPs and noise values.
% plot the win loss rate or something

\begin{figure}[t!]
    \centering
    \begin{minipage}[b]{0.32\textwidth}
        \centering
        \includegraphics[width=\textwidth]{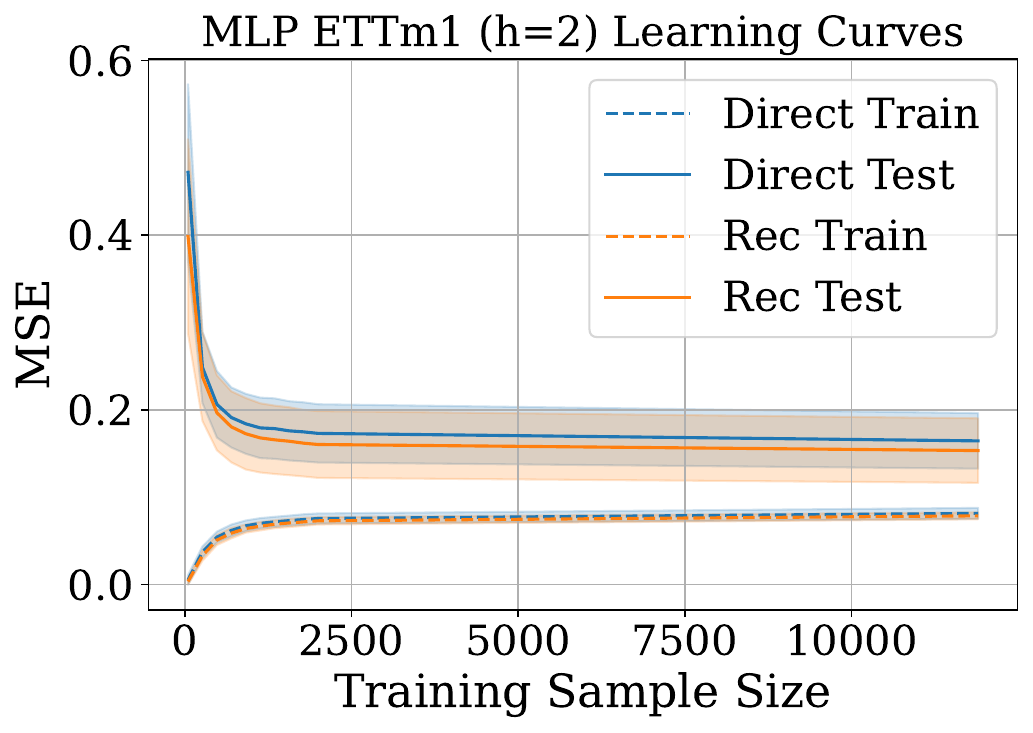}
    \end{minipage}
    \hfill
    \begin{minipage}[b]{0.32\textwidth}
        \centering
        \includegraphics[width=\textwidth]{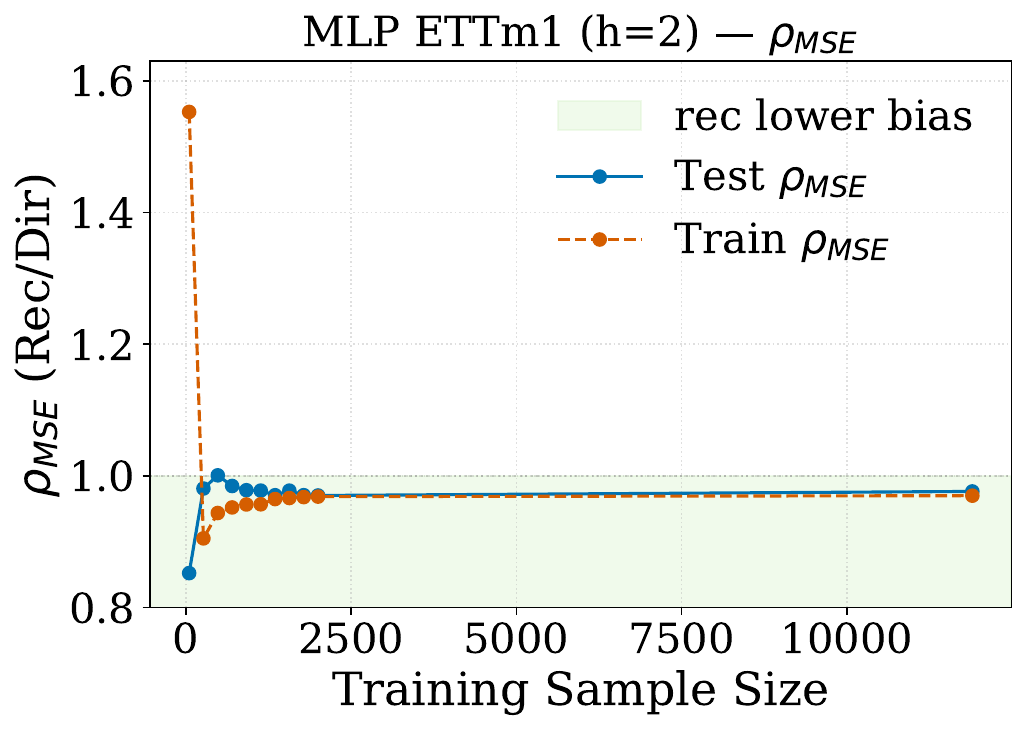}
    \end{minipage}
    \hfill
    \begin{minipage}[b]{0.32\textwidth}
        \centering
        \includegraphics[width=\textwidth]{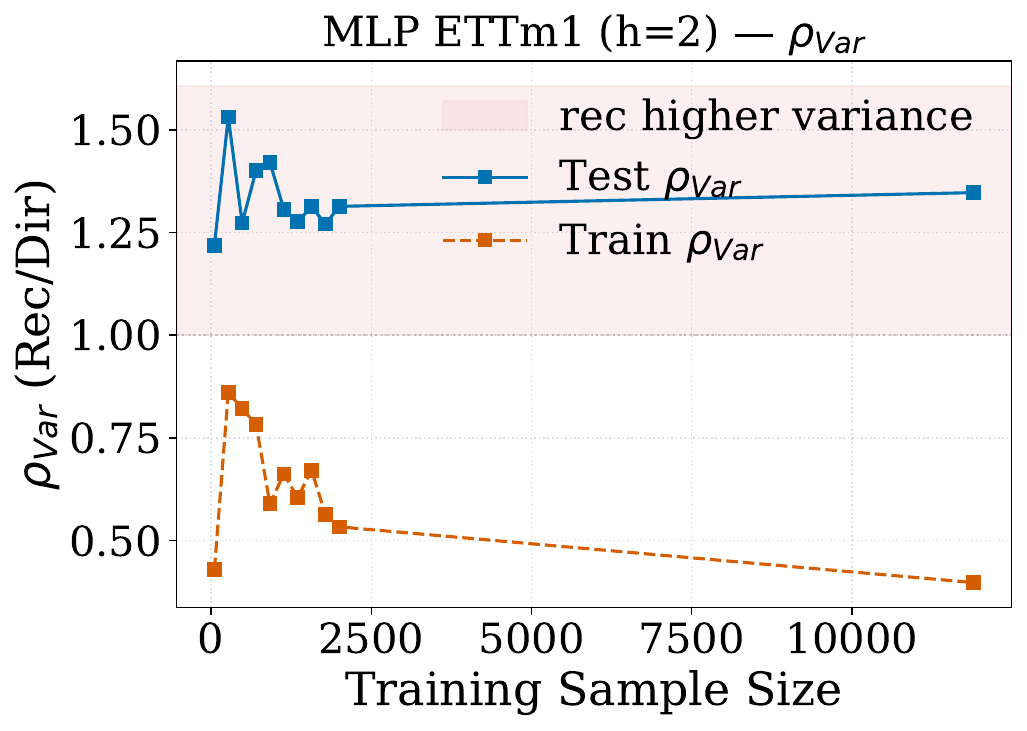}
    \end{minipage}
    \caption{
\textbf{MLP forecasting on ETTm1 ($h{=}2$): comparison of learning curves and ratio-based metrics.}
Each panel compares \emph{recursive} and \emph{direct} MLPs trained under identical capacity and data conditions (lags$=50$, width$=2$, 50 seeds). 
(left)~Learning curves show effective learning and capacity saturation. 
(middle)~$\rho_{MSE}$ is rhe relative MSE and (right)~~$\rho_{VAR}$ denote the \emph{ratios} of recursive to direct errors ($\mathrm{Rec}/\mathrm{Dir}$), where $y{=}1$ indicates parity.
Across sample sizes, the recursive model exhibits lower bias (train/test ~$\rho_{MSE}$~$<1$ at large~$N$) but higher estimation variance (~$\rho_{VAR}$~$>1$), 
supporting our theoretical findings that recursion is not necessarily higher bias and can amplify variance.
}

    \label{fig:mlp_learning_curves}
\end{figure}

\paragraph{Empirical validation on real data.}
To complement our synthetic analyses, we evaluate the direct and recursive strategies on the commonly used \textbf{ETTm1} dataset \cite{zhou2021informer}. 
We fix the forecast horizon to $h{=}2$ and use an input window of 50 lags, providing the recursive model with sufficient expressivity, as suggested by our earlier analysis. 
Both strategies employ a shallow MLP with hidden width $2$, deliberately chosen to \emph{bottleneck model capacity} rather than data availability. 
We repeat each experiment $50$ times over random seeds to obtain stable mean and variance estimates.

Across all runs, the recursive model achieves consistently lower training and test error. The plateau on the left of Figure \ref{fig:mlp_learning_curves} shows we reach the capacity bottleneck.
Test MSE variances are higher for the recursive model at all sample sizes, signalling higher estimation variance. 
These observations align closely with our theoretical findings: the recursive strategy can exhibit \emph{higher variance} and not necessarily higher bias.

% \paragraph{Practical takeaway.}
% Our results challenge the common simplification that ``recursion is more biased.'' 
% Both our theoretical and empirical analyses show that this statement does not hold universally and may explain why, in practice, large performance gaps and uncertainty persist between direct and recursive forecasting strategies. 
% By clarifying when recursion benefits or suffers, our study provides a more grounded understanding of how model capacity, data regime, and compositional dynamics interact. 
% % This observation aligns with broader evidence that simpler linear architectures can rival or surpass Transformer-based models on time-series forecasting tasks \cite{zeng2023transformers}, reinforcing that structural and variance trade-offs—rather than architectural complexity alone—often determine performance.
% % add aout linear beating rtansformer as telmo said

% \paragraph{Limitations and future work.}
% While our experiments demonstrate a lower effective bias for the recursive MLP, we have not yet verified whether this behaviour stems from the same mechanism identified in our earlier analytic study—namely, the Jacobian of the composition function. 
% Validating this connection for nonlinear neural predictors is an important direction for future work. Furthermore, we aim to use this new theoretical perspective to guide selection of a strategy, as a prior for HP-optimisation, as suggested by recent studies \cite{Green2025Stratify}.
% % reference

\paragraph{Practical implications and limitations.}
Our results show that recursion is not inherently more biased: depending on the data regime and model nonlinearity, recursive predictors can exhibit \emph{lower bias but higher estimation variance} than direct ones. This challenges the idea that greater architectural complexity automatically yields better forecasts. In fact, as recent studies suggest \citep{zeng2023transformers}, well-tuned linear models can match or surpass Transformer-based forecasters. Our analysis sits at this junction: by formalising how recursive composition amplifies estimation variance yet can reduce structural bias, we extend the explanatory scope of linear-model behaviour toward nonlinear predictors.
While our experiments confirm the predicted bias–variance reversal, they do not yet isolate an exact form for the Jacobian-based mechanism for nonlinear predictors. Future work should quantify this link by estimating composition-Jacobian norms across architectures and horizons—and extend the analysis to deeper or more expressive models. Leveraging these diagnostics as priors for automatic strategy or hyperparameter selection, as proposed by \cite{Green2025Stratify}, represents a promising next step.

\section{Conclusion}
We revisited the bias–variance trade-off in multi-step forecasting through an epistemic error decomposition separating irreducible noise, structural bias, and estimation variance. Our analysis shows that recursive composition reshapes model expressivity and uncertainty propagation: recursion can reduce bias and/or amplify estimation variance, depending on data regime and nonlinearity. This framework unifies linear and nonlinear forecasting behaviour under a single geometric view and provides a principled basis for diagnosing when each strategy should dominate. Finally, by distinguishing the roles of process and measurement noise, we show that the ``best'' strategy is domain-contingent—epistemic uncertainty is not simply higher or lower, but structured by the interaction between data dynamics, model capacity, and compositional geometry. Future work will extend this analysis to richer architectures and empirically quantify the Jacobian-driven amplification of uncertainty, bridging theoretical insight with practical model selection.
% We decomposed multi-step forecasting error into three epistemic components: irreducible noise, the structural gap $\Ggap$, and the estimation variance $\EV$. 
% This view reframes the long-standing bias--variance debate between recursive and direct strategies as a question of how uncertainty is generated, propagated, and amplified through composition. 
% For linear predictors, $G_{rec} = G_{dir}$, but nonlinear composition can invert this ordering, making bias data-dependent rather than strategy-determined. 
% The recursive estimation variance follows a Jacobian-driven amplification $\Th$, offering a measurable diagnostic of epistemic uncertainty under recursion. 
% Finally, distinguishing the roles of process and measurement noise reveals that the ``best'' strategy is domain-contingent: epistemic uncertainty is not merely higher or lower, but structured by the interaction between data dynamics, model capacity, and compositional geometry.

\newpage
% =======================
% References
% =======================
\bibliographystyle{plainnat}
\bibliography{references}

\appendix

\newpage

\section{Appendix: Notation Table}
\label{notation}
% \label{app:notation}

\begin{table}[h!]
\centering
\caption{Summary of notation used throughout the paper.}
\renewcommand{\arraystretch}{1.15} % Adjust row spacing for readability
\setlength{\tabcolsep}{8pt}      % Adjust column spacing
\begin{tabular}{ll}
\toprule
\textbf{Symbol} & \textbf{Definition} \\
\midrule
\multicolumn{2}{l}{\textit{General Mathematical Operators}} \\
$\E[\cdot]$ & Expectation operator. \\
$\mathbb{R}$ & The set of real numbers. \\
$\bm{v}$ & Generic notation for a vector (e.g., $\x, \w, \thetaV$). \\
\midrule

\multicolumn{2}{l}{\textit{Time Series and Data Generating Process (DGP)}} \\
$y_t$ & Observed time series value at time $t$. \\
$\{x_t\}$ & The stationary latent (unobserved) process. \\
$v_t$ & Measurement noise ($y_t = x_t + v_t$). \\
$w_t$ & Process noise in the latent dynamics. \\
$h$ & Forecast horizon. \\
$p$ & Number of lagged observations used as input. \\
$N$ & Number of training samples. \\
$\sigma_s^2, \sigma_e^2$ & Variance of process and measurement noise, respectively. \\
$a, \gamma$ & Parameters of the latent AR(2) process in noise analysis. \\
\midrule

\multicolumn{2}{l}{\textit{Forecasting Strategies \& Predictors}} \\
$\Rec$ & The recursive forecasting strategy. \\
$\Dir$ & The direct forecasting strategy. \\
$f_h^\star$ & The Bayes-optimal $h$-step predictor. \\
$f_h^{\mathcal H}$ & The best-in-class oracle predictor for hypothesis space $\mathcal{H}$. \\
$\hat{f}_h^{\mathcal H}$ & The finite-sample estimator trained on $N$ samples. \\
\midrule

\multicolumn{2}{l}{\textit{Model Parameters \& Mappings}} \\
$\thetaV$ & Vector of one-step model parameters (e.g., $[b_1, b_2]$). \\
$\hat{\thetaV}$ & A finite-sample estimator of $\thetaV$. \\
$\alpha_h$ & Vector of effective $h$-step parameters for the recursive model. \\
$c_h$ & Vector of parameters for the direct $h$-step model. \\
$\Tmap_h$ & The parameter composition map, $\Tmap_h: \thetaV \mapsto \alpha_h(\thetaV)$. \\
$A$ & A specific instance of $\Tmap_h$ used for $h=2$ in examples. \\
\midrule

\multicolumn{2}{l}{\textit{Error Decomposition \& Analysis}} \\
$\MSE$ & Mean Squared Error. \\
$\sigma^2_{\text{noise}}$ & Irreducible aleatoric noise variance. \\
$G_{\mathcal H}$ & Structural gap (approximation bias) for class $\mathcal{H}$. \\
$\EV_{\mathcal H}$ & Estimation variance for class $\mathcal{H}$. \\
$\Ggap$ & \textbf{Difference} in structural gaps: $G_{\Rec} - G_{\Dir}$. \\
$EV_{\Delta}$ & \textbf{Difference} in estimation variances: $EV_{\Rec} - EV_{\Dir}$. \\
$\Th$ & Scalar amplification factor for recursive $\EV$ (see Theorem 1). \\
$\rho_{\text{MSE}}, \rho_{\text{VAR}}$ & \textbf{Ratio} of errors (Rec/Dir) used in empirical plots. \\
\midrule

\multicolumn{2}{l}{\textit{Feature Vectors \& Geometric Objects}} \\
$\x_t$ & Vector of $p$ lagged features, $[y_t, \dots, y_{t-p+1}]$. \\
$\tilde{\x}_t$ & Effective feature vector induced by $h$-step recursive composition. \\
$Q$ & Second-moment matrix of base features: $\E[\x_t \x_t^\top]$. \\
$\widetilde{Q}$ & Second-moment matrix of composed features: $\E[\tilde{\x}_t \tilde{\x}_t^\top]$. \\
$\Jh$ & Jacobian matrix of the composition map $\Tmap_h$. \\
$\Sigma_\theta$ & Covariance matrix of the one-step parameter estimator $\hat{\thetaV}$. \\
\bottomrule
\end{tabular}
\label{tab:notation}
\end{table}

\newpage
\section{Appendix: Proof Details}
\label{app:proofs}
\subsection{Proof that linear predictors are surjective on $\mathbb{R}^2$:}
\label{surjective_proof}

We want to prove that for every $(\phi_{1}, \phi_{2}) \in \mathbb{R}^2$, there exists at least one $(b_{1}, b_{2}) \in \mathbb{R}^2$ such that $A(b_{1}, b_{2}) = (b_{1}^2 + b_{2}, b_{1}b_{2}) = (\phi_{1}, \phi_{2})$. 

Rearrange $\phi_{1} = b_{1}^2 + b_{2}$ to instead write: $b_{2} = \phi_{1}-b_{1}^2$ and substitute into: 

$\phi_{2} = b_{1}b_{2} = b_{1} (\phi_{1} - b_{1}^2) = b_{1}\phi_{1} - b_{1}^3$. This can be rearranged to be a cubic polynomial in $b_{1}$: $b_{1}^3 - \phi_{1}b_{1} + \phi_{2}$ = 0. 

Given that this is a cubic equation with real coefficients, there exists at least one real root to this equation (since any complex roots must come as a conjugate pair in order for the equation to have real coefficients). We can call this real root $x$ such that $x$ is real and $x^3 - \phi_{1}x+\phi_{2} = 0$. 

Now we simply substitute $x$ into our equation for $\phi_{1}$. Write our candidate value for $b_{2}$ as $y$ so that we have: $y = \phi_{1} - x$. 

Therefore for every $(\phi_{1}, \phi_{2}) \in \mathbb{R}^2$ there exists $(x, y) \in \mathbb{R}^2$, as defined above, such that $A(x, y) = (\phi_{1}, \alpha_{2}).$

% \newpage
\subsection{Linearisation proof}
\label{jacob_proof}

\paragraph{Setup and notation.}
Let $X\in\mathbb{R}^d$ be a random vector with finite second moments,
mean $\mu \coloneqq \mathbb{E}[X]$, and covariance
$\Sigma \coloneqq \Cov(X)
= \mathbb{E}\!\left[(X-\mu)(X-\mu)^\top\right]$.
Let $\alpha:\mathbb{R}^d\to\mathbb{R}$ be differentiable in a
neighbourhood of $\mu$, and write its gradient at $\mu$ as
$g \coloneqq \nabla \alpha(\mu) \in \mathbb{R}^d$.
For a scalar random variable $Z$, we use
$\operatorname{Var}(Z)=\mathbb{E}\!\left[(Z-\mathbb{E}Z)^2\right]$;
for vectors we use $\Cov(\cdot)$ as above.

\paragraph{Claim (delta-method linearisation).}
\[
\operatorname{Var}\!\big(\alpha(X)\big)
\;=\; g^\top \Sigma\, g \;+\; o(\|\Sigma\|),
\qquad\text{hence}\qquad
\operatorname{Var}\!\big(\alpha(X)\big)\;\approx\; g^\top \Sigma\, g .
\]

\paragraph{Proof.}
A first-order Taylor expansion of $\alpha$ at $\mu$ gives
\begin{equation*}
\alpha(X) \;=\; \alpha(\mu) + g^\top (X-\mu) + R(X),
\end{equation*}
where the remainder $R(X)=o(\|X-\mu\|)$ as $X\to\mu$ (e.g.\ if $\alpha$ is
$C^1$) and, with $C^2$, one may write
$R(X)=\tfrac12 (X-\mu)^\top H(\xi)(X-\mu)$ for some $\xi$ on the line
segment between $X$ and $\mu$ ($H$ is the Hessian).
Since $\alpha(\mu)$ is constant and $\mathbb{E}[X-\mu]=0$,
\begin{align*}
\operatorname{Var}\!\big(\alpha(X)\big)
&= \operatorname{Var}\!\big(g^\top (X-\mu) + R(X)\big) \\
&= \operatorname{Var}\!\big(g^\top (X-\mu)\big)
   \;+\; 2\,\Cov\!\big(g^\top (X-\mu),\, R(X)\big)
   \;+\; \operatorname{Var}\!\big(R(X)\big).
\end{align*}
Because $R(X)=o(\|X-\mu\|)$, the last two terms are $o(\|\Sigma\|)$ whenever
$\mathbb{E}\|X-\mu\|^2<\infty$; hence, to first order,
\[
\operatorname{Var}\!\big(\alpha(X)\big)
\;=\; \mathbb{E}\!\left[(g^\top(X-\mu))^2\right] + o(\|\Sigma\|)
\;=\; g^\top\,\mathbb{E}\!\left[(X-\mu)(X-\mu)^\top\right] g + o(\|\Sigma\|)
\;=\; g^\top \Sigma\, g + o(\|\Sigma\|).
\]
\hfill$\square$

\paragraph{Vector-output extension.}
If $\alpha:\mathbb{R}^d\to\mathbb{R}^k$: 

\[\alpha(\mathbf{x}) = (\alpha_{1}(\mathbf{x}), \ldots, \alpha_{k}(\mathbf{x}))^{\top}\] 
with Jacobian matrix $J_{\alpha} \in \mathbb{R}^{k \times d}$ defined by:

\[[J_{\alpha}(y)]_{ij} = \frac{\partial \alpha_{i}}{\partial x_{j}}(y).\]
%$J\coloneqq \nabla \alpha(\mu)\in\mathbb{R}^{k\times d}$

A first order Taylor series expansion yields:

\[\alpha(\mathbf{x}) \approx \alpha(\mathbf{\mu}) + J_{\alpha}(\mu)(\mathbf{x - \mu)}.\]

Using the definition of Covariance for a random vector:

\[\Cov(\alpha(\mathbf{x})) = \mathbb{E}[(\alpha(\mathbf{x})-\mathbb{E}[\alpha(\mathbf{x})]) (\alpha(\mathbf{x})-\mathbb{E}[\alpha(\mathbf{x})])^{\top}]\]

and substituting in \(\alpha(\mathbf{x}) - \mathbb{E}[\alpha(\mathbf{x})] \approx  J_{\alpha}(\mu)(\mathbf{x - \mu)}\), the same argument yields:

\[\Cov(\alpha(\mathbf{x})) \approx \mathbb{E}[J_{\alpha}(\mu) \mathbf{(x - \mu)(x-\mu)^{\top}}J_{\alpha}(\mu)^{\top}]\]
\[=J_{\alpha}(\mu) \mathbb{E}[\mathbf{(x - \mu)(x-\mu)^{\top}}] J_{\alpha}(\mu)^{\top}\]
\[=J_{\alpha}(\mu) \Cov(\mathbf{x}) J_{\alpha}(\mu)^{\top}.\]

To conclude we have:
\[
\Cov\!\big(\alpha(X)\big)
\;=\; J\,\Sigma\,J^\top \;+\; o(\|\Sigma\|),
\quad\text{so}\quad
\Cov\!\big(\alpha(X)\big)\approx J\,\Sigma\,J^\top .
\]

% \newpage

\subsection{EV Proof}
\label{ev_proof}

\paragraph{Base vs composed feature geometry.}
Let $z_t$ denote the information set at time $t$ (e.g., $p$ lags). 
The \emph{base} one–step (and direct) feature map is $\phi:\mathcal{Z}\!\to\!\mathbb{R}^{p}$ with features
\[
x \;=\; \phi(z_t), 
\qquad 
Q \;:=\; \mathbb{E}[\,x\,x^\top\,].
\]
Under recursive $h$–step composition, the \emph{effective} $h$–ahead predictor remains linear-in-parameters but acts on a (generally richer) \emph{composed} feature map $\phi_h$.
We write
\[
\tilde{x} \;=\; \phi_h(z_t),
\qquad 
\widetilde{Q} \;:=\; \mathbb{E}[\,\tilde{x}\,\tilde{x}^\top\,].
\]

\paragraph{Parameterisation and covariances.}
Let $\theta\!\in\!\mathbb{R}^d$ be the one–step parameter vector with estimator $\hat\theta$ and covariance 
$\Sigma_\theta = \mathrm{Cov}(\hat\theta)$.
Let $\alpha_h = g_h(\theta)$ be the effective $h$–ahead parameters obtained by composing the one–step model $h$ times; let $J_h := \frac{\partial g_h}{\partial \theta}(\theta^\star)$ be the Jacobian at the population optimum $\theta^\star$.

\begin{lemma}[Delta method for the composition map]
If $g_h$ is differentiable in a neighbourhood of $\theta^\star$, then
\[
\Sigma_{\alpha_h} 
\;\approx\; J_h\,\Sigma_\theta\,J_h^\top.
\]
\end{lemma}

\begin{proposition}[EV for recursion]
Assume (i) linear-in-parameters predictors at the horizon, (ii) exogenous test features independent of estimation noise, and (iii) the delta-method approximation above. Then the finite-sample estimation-variance term of the recursive strategy at horizon $h$ is
\[
\boxed{\quad
\mathrm{EV}_{\mathrm{rec}}(h) 
\;\approx\; \mathrm{tr}\!\big( J_h\,\Sigma_\theta\,J_h^\top\,\widetilde{Q} \big).
\quad}
\]
\end{proposition}

\paragraph{One–step baseline and the ratio \(T_h\).}
The corresponding one–step EV uses the \emph{same} base geometry $Q$:
\[
\mathrm{EV}_{1} \;=\; \mathrm{tr}\!\big(\Sigma_\theta\,Q\big).
\]
Define the (dimensionless) amplification factor
\[
\boxed{\quad
T_h 
\;:=\; 
\frac{\mathrm{tr}\!\big(J_h\,\Sigma_\theta\,J_h^\top\,\widetilde{Q}\big)}
     {\mathrm{tr}\!\big(\Sigma_\theta\,Q\big)}
\quad}
\qquad\Longrightarrow\qquad
\mathrm{EV}_{\mathrm{rec}}(h)\;\approx\; T_h\,\mathrm{EV}_{1}.
\]

\paragraph{Direct strategy.}
If the direct $h$–ahead model uses the \emph{same} base features $\phi$, its EV reads
\[
\boxed{\quad
\mathrm{EV}_{\mathrm{dir}}(h) \;\approx\; \mathrm{tr}\!\big(\Sigma_\theta^{(h)}\,Q\big),
\quad}
\]
where $\Sigma_\theta^{(h)}$ is the estimator covariance of the direct $h$–ahead parameters.
(Under well-specified OLS with i.i.d.\ noise, $\Sigma_\theta^{(h)} \approx \tfrac{\sigma^2_{\varepsilon,h}}{N}\,Q^{-1}$, hence $\mathrm{EV}_{\mathrm{dir}}(h) \approx \tfrac{\sigma^2_{\varepsilon,h}}{N}\,p$.) 

\paragraph{EV difference.}
Recall $Q$ is shared by one–step and direct, while recursion induces $\widetilde{Q}$,
\[
\boxed{\quad
\Delta \mathrm{EV}(h) 
\;:=\; \mathrm{EV}_{\mathrm{rec}}(h) - \mathrm{EV}_{\mathrm{dir}}(h)
\;\approx\; \mathrm{tr}\!\big(J_h\,\Sigma_\theta\,J_h^\top\,\widetilde{Q}\big)
         \;-\; \mathrm{tr}\!\big(\Sigma_\theta^{(h)}\,Q\big).
\quad}
\]

\textbf{Note on nonlinear predictors (e.g., MLPs).}
\label{mlp_note}
For differentiable predictors that are non-linear in parameters (e.g., MLPs), the above EV expressions admit a first-order (delta-method) extension by replacing $Q$ and $\widetilde{Q}$ with parameter–Jacobian Gram matrices. A full derivation is left to future work.

% \newpage
\subsection{Aleatoric noise for $\sigma_{\varepsilon,1}$ and $\sigma_{\varepsilon,2}$}
\label{noise_proof}

\paragraph{Setup.}
Let the latent AR(2) state evolve as
\begin{equation}
x_t \;=\; a x_{t-1} + \gamma x_{t-2} + w_t, 
\qquad w_t \sim \mathcal N(0,\sigma_s^2),
\end{equation}
and the observations be
\begin{equation}
y_t \;=\; x_t + v_t, 
\qquad v_t \sim \mathcal N(0,\sigma_e^2),
\end{equation}
with $\{w_t\}$ and $\{v_t\}$ mutually independent and independent over time. 
We consider oracle AR predictors that use the true $(a,\gamma)$ on the observed series $y$.
Define the $h$-step composed coefficients by 
$c^{(h)} = \big(c^{(h)}_1,c^{(h)}_2\big)$ so that
\begin{equation}
\hat y_{t+h} \;=\; c^{(h)}_1\,y_t + c^{(h)}_2\,y_{t-1}.
\end{equation}
For AR(2),
\begin{equation}
c^{(1)}=(a,\gamma), 
\qquad
c^{(2)}=\big(a^2+\gamma,\; a\gamma\big).
\end{equation}

\paragraph{Claim.}
For linear predictor, $\Ggap = 0$, so the irreducible (aleatoric) mean-squared errors at horizons $h=1,2$ are
\begin{align}
\sigma_{\varepsilon,1}^2 
&:= \E\big[(y_{t+1}-a y_t - \gamma y_{t-1})^2\big]
\;=\; \sigma_s^2 \;+\; \big(1+a^2+\gamma^2\big)\sigma_e^2, \label{eq:aleatoric-1} \\
\sigma_{\varepsilon,2}^2 
&:= \E\big[(y_{t+2}-(a^2+\gamma)y_t - (a\gamma) y_{t-1})^2\big]
\;=\; (1+a^2)\sigma_s^2 \;+\; \Big(1+(a^2+\gamma)^2 + (a\gamma)^2\Big)\sigma_e^2. \label{eq:aleatoric-2}
\end{align}

\paragraph{Proof.}
Write $y_t=x_t+v_t$ and expand the $h$-step prediction error $e_{t+h}:=y_{t+h}-\hat y_{t+h}$ into a process-noise part (from future $w$'s) and a measurement-noise part (from $v$'s).

\emph{(i) One step.}
Using $x_{t+1}=a x_t+\gamma x_{t-1}+w_{t+1}$ and $\hat y_{t+1}=a y_t+\gamma y_{t-1}$,
\begin{align*}
e_{t+1}
&= (x_{t+1}+v_{t+1}) - \big(a(x_t+v_t)+\gamma(x_{t-1}+v_{t-1})\big) \\
&= \big(x_{t+1}-ax_{t} - \gamma x_{t-1} \big) + \big(v_{t+1}-a v_t-\gamma v_{t-1}\big) \\
&= \underbrace{w_{t+1}}_{\text{process error}} 
\;+\; \underbrace{\big(v_{t+1}-a v_t-\gamma v_{t-1}\big)}_{\text{measurement error}}.
\end{align*}

It is known that $\Var(aX+bY)=a^2\Var(X)+b^2\Var(Y) + 2ab \Cov(X, Y)$ and we know that cross terms are independent and so the covariance term vanishes, therefore we have:
\[\Var(e_{t+1})=\Var\big(w_{t+1}+\big(v_{t+1}-a v_t-\gamma v_{t-1}\big)\big)\]
\[= \Var(w_{t+1}) + \Var(v_{t+1}) + a^2 \Var(v_t) + \gamma^2 \Var(v_{t-1})
= \sigma_s^2 + (1+a^2+\gamma^2)\sigma_e^2,
\]

\emph{(ii) Two steps.}
From $x_{t+2}=a x_{t+1}+\gamma x_t + w_{t+2}$ and $x_{t+1}=a x_t + \gamma x_{t-1}+w_{t+1}$ we get
\[
x_{t+2}=(a^2+\gamma) x_t + (a\gamma) x_{t-1} + a w_{t+1} + w_{t+2}.
\]
With $\hat y_{t+2}=(a^2+\gamma) y_t + (a\gamma) y_{t-1}$,
\begin{align*}
e_{t+2}
&= (x_{t+2}+v_{t+2}) - \big((a^2+\gamma)(x_t+v_t)+(a\gamma)(x_{t-1}+v_{t-1})\big) \\
&= \underbrace{(a w_{t+1}+w_{t+2})}_{\text{process}}
\;+\; \underbrace{\Big(v_{t+2}-(a^2+\gamma)v_t-(a\gamma)v_{t-1}\Big)}_{\text{measurement}}.
\end{align*}
Again, cross-terms vanish by independence, so
\[
\Var[e_{t+2}]
= (a^2+1)\sigma_s^2 \;+\; \Big(1+(a^2+\gamma)^2+(a\gamma)^2\Big)\sigma_e^2,
\]
which is \eqref{eq:aleatoric-2}. \qed

\paragraph{Discussion.}
Equations \eqref{eq:aleatoric-1}--\eqref{eq:aleatoric-2} are the aleatoric floors achieved by oracle predictors on $y$. 
They show that, depending on $(a,\gamma,\sigma_s,\sigma_e)$, it is possible for $\sigma_{\varepsilon,2}^2 < \sigma_{\varepsilon,1}^2$, i.e., the two-step observation can be \emph{intrinsically less noisy} than the one-step. 
This asymmetry explains why recursive training (which chains noisy one-step mappings) may incur higher estimation variance than direct multi-step training.
% \newpage

\subsection{Special case linear predictor linear DGP}
Under homoskedastic (our sequence has constant and finite variance), well-specified Ordinary Least Squares (OLS) regression with i.i.d.\ noise at horizon $h$, 
\[
  \Sigma_{\theta_h} \;=\; \frac{\sigma_{\varepsilon,h}^2}{N}\,Q_h^{-1}
  \qquad\Longrightarrow\qquad
  \EV_{\Dir}(h) \;=\; \frac{\sigma_{\varepsilon,h}^2}{N}\,\text{trace}(I_{p_h})
  \;=\; \frac{\sigma_{\varepsilon,h}^2}{N}\,p_h.
\]

% \newpage

\section{Appendix: Figure \ref{fig:bilinear_G_comp} Experimental Set Up
}
\label{fig_1_details}
We compare a direct two–step bilinear predictor with a recursively composed one–step bilinear predictor, both evaluated inside a common 6-dimensional polynomial coefficient space.

\textbf{Task space.}
Let
\[
\psi(y_t, y_{t-1})
=\big[y_t,\; y_{t-1},\; y_t y_{t-1},\; y_t^2,\; y_t^2 y_{t-1},\; y_{t-1}^2\big]^\top,
\]
and represent any two–step polynomial predictor as
\[
\hat y_{t+2} = \theta^\top \psi(y_t, y_{t-1}), \qquad
\theta = (\theta_1, \dots, \theta_6) \in \mathbb{R}^6.
\]
This defines the ambient \emph{task space} $\Theta \subset \mathbb{R}^6$.

\textbf{Model families.}
The direct two–step bilinear predictor uses three basis terms:
\[
\hat y^{\text{Dir}}_{t+2} = c_1 y_t + c_2 y_{t-1} + c_3 y_t y_{t-1}, \qquad c \in \mathbb{R}^3.
\]
This forms a 3D linear subspace 
$\mathcal{F}_c = \mathrm{span}\{y_t, y_{t-1}, y_t y_{t-1}\} \subset \mathbb{R}^6$.

The recursive predictor is obtained by composing a one–step bilinear mapping:
\[
\hat y_{t+1} = b_1 y_t + b_2 y_{t-1} + b_3\, y_t y_{t-1}, \qquad b \in \mathbb{R}^3,
\]
\[
\hat y_{t+2} = b_1 \hat y_{t+1} + b_2 y_t + b_3\, \hat y_{t+1} y_t.
\]
Expanding and collecting terms yields the composed two–step predictor:
\[
\boxed{
\hat y_{t+2}
= \alpha_1 y_t
+ \alpha_2 y_{t-1}
+ \alpha_3 y_t y_{t-1}
+ \alpha_4 y_t^2
+ \alpha_5 y_t^2 y_{t-1},
}
\]
with coefficients determined by the composition map $\alpha = g(b)$:
\[
\boxed{
\alpha_1 = b_1 + b_2^{2}, \qquad
\alpha_2 = 2 b_1 b_2, \qquad
\alpha_3 = b_3 (b_1 + b_2), \qquad
\alpha_4 = b_3 b_1, \qquad
\alpha_5 = b_3^{2}.
}
\]
Hence, the recursive family forms a nonlinear 3D manifold
$\mathcal{F}_\alpha = g(\mathbb{R}^3) \subset \mathbb{R}^5 \subset \mathbb{R}^6$,
while the direct family $\mathcal{F}_c$ remains a 3D linear subspace.

\textbf{Sampling protocol.}
We uniformly sample base parameters
\(
b \sim \mathrm{Unif}([-1.5, 1.5]^3)
\)
and map them through $g$ to obtain a cloud of composed coefficients $\alpha$.
Let $[a_i^-, a_i^+]$ denote the coordinatewise bounds of this cloud for $i=1,\dots,5$.
These bounds define the support for both the direct and recursive families:
\[
\mathcal{A} = \prod_{i=1}^{5} [a_i^-, a_i^+] \subset \mathbb{R}^5,
\qquad
\mathcal{C} = \prod_{i=1}^{3} [a_i^-, a_i^+] \subset \mathbb{R}^3.
\]
The full six–dimensional task box is then
\[
\Theta = \mathcal{A} \times [-1.5, 1.5] \subset \mathbb{R}^6,
\]
where the final coordinate $\theta_6$ multiplies the term $y_{t-1}^2$, which
neither model can represent.

\textbf{Evaluation.}
Tasks $\theta \sim \mathrm{Unif}(\Theta)$ are drawn at random.
For each task, we simulate data from the corresponding generative model
$y_{t+2} = \theta^\top \psi(y_t, y_{t-1})$
and fit both predictors under identical training and evaluation conditions.
We compute:
(i) the \emph{coefficient–structure distance} to each model family,
\[
d_\alpha(\theta) = \min_{\alpha \in \mathcal{F}_\alpha} \|P_5 \theta - \alpha\|_2,
\qquad
d_c(\theta) = \min_{c \in \mathbb{R}^3} \|P_3 \theta - (c_1, c_2, c_3)\|_2,
\]
where $P_k$ projects onto the first $k$ coordinates; and
(ii) the test MSEs, $\mathrm{MSE}_\alpha$ and $\mathrm{MSE}_c$.
ECDFs of $(d_\alpha, d_c)$ quantify structural bias,
while pairwise $\mathrm{MSE}_\alpha$–vs–$\mathrm{MSE}_c$ plots show behavioural bias.

\section{Appendix: Figure \ref{fig:sigma_e} Experimental Set Up
}
\label{fig_3_details}
To produce the results in Figure~\ref{fig:sigma_e}, we designed a Monte Carlo simulation to empirically measure the estimation variance (EV) and test the validity of our theoretical model under varying noise conditions.

\paragraph{Data Generating Process (DGP)}
The data is generated from a linear latent AR(2) process with additive measurement noise. The latent state $x_t$ evolves according to $x_t = a x_{t-1} + \gamma x_{t-2} + w_t$, where $w_t$ is i.i.d. process noise drawn from $\mathcal{N}(0, \sigma_s^2)$. The observations $y_t$ are given by $y_t = x_t + v_t$, where $v_t$ is i.i.d. measurement noise from $\mathcal{N}(0, \sigma_e^2)$. For the experiment, we fixed the process noise at $\sigma_s = 1$ and swept the DGP parameters $a$ and $\gamma$ across their stable regions. The process and measurement noise was varied from 0 to 1.0 to study its impact.

\paragraph{Models and Training}
We used a linear predictor with two lagged features ($y_t$, $y_{t-1}$) for all strategies.
\begin{itemize}
    \item The \textbf{one-step} and \textbf{direct ($h=2$)} predictors were trained independently on each simulated dataset using Ordinary Least Squares (OLS) without an intercept.
    \item The \textbf{recursive ($h=2$)} predictor's coefficients were not trained directly, but were derived by applying the nonlinear composition map ($[b_1, b_2] \to [b_1^2 + b_2, b_1 b_2]$) to the coefficients of the fitted one-step model.
\end{itemize}

\paragraph{Evaluation Protocol}
We ran 50 trials (seeds) for each unique parameter configuration ($a, \gamma, \sigma_s, \sigma_e$). In each trial, a new training set of $N_{\text{tr}} = 10000$ samples was generated. All models were then evaluated on a large, fixed test set of $N_{\text{ev}} = 20000$ samples drawn from the same DGP. The empirical EV was calculated as the mean variance of the predictions across all trials for each point in the test set, i.e., $\text{EV}_{\text{emp}} = \mathbb{E}_{x_{\text{test}}}[\text{Var}_{\text{trials}}(\hat{f}(x_{\text{test}}))]$. This provides a stable, finite-sample measurement of the estimation variance to compare against our theoretical model.

\paragraph{MSE of Recursive and Direct under $a, \gamma,  \sigma_e,  \sigma_s$ sweeps}
For the sweeps in Figure \ref{fig:sigma_e}, we report the empirical MSE of the recursive and direct strategy and compare it to the theoretical aleatoric noise from Equations \ref{eq:aleatoric-1} and \ref{eq:aleatoric-2}.

\begin{figure}[htbp]
    \centering
    % Left: Analytic
    \begin{subfigure}[t]{0.48\textwidth}
        \centering
        \includegraphics[width=\textwidth]{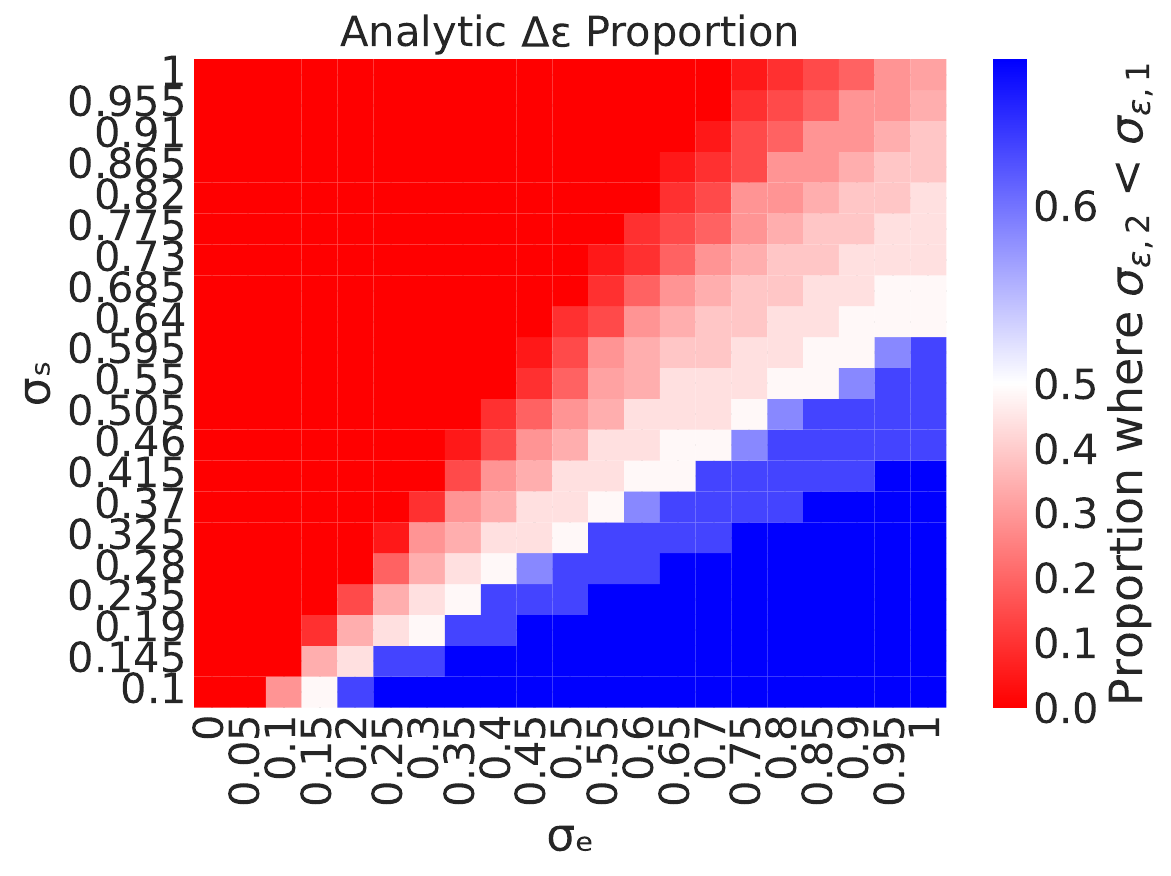}
        \caption{Analytic $\Delta \varepsilon$ Proportion}
        \label{fig:analytic_prop}
    \end{subfigure}
    \hfill
    % Right: Empirical
    \begin{subfigure}[t]{0.48\textwidth}
        \centering
        \includegraphics[width=\textwidth]{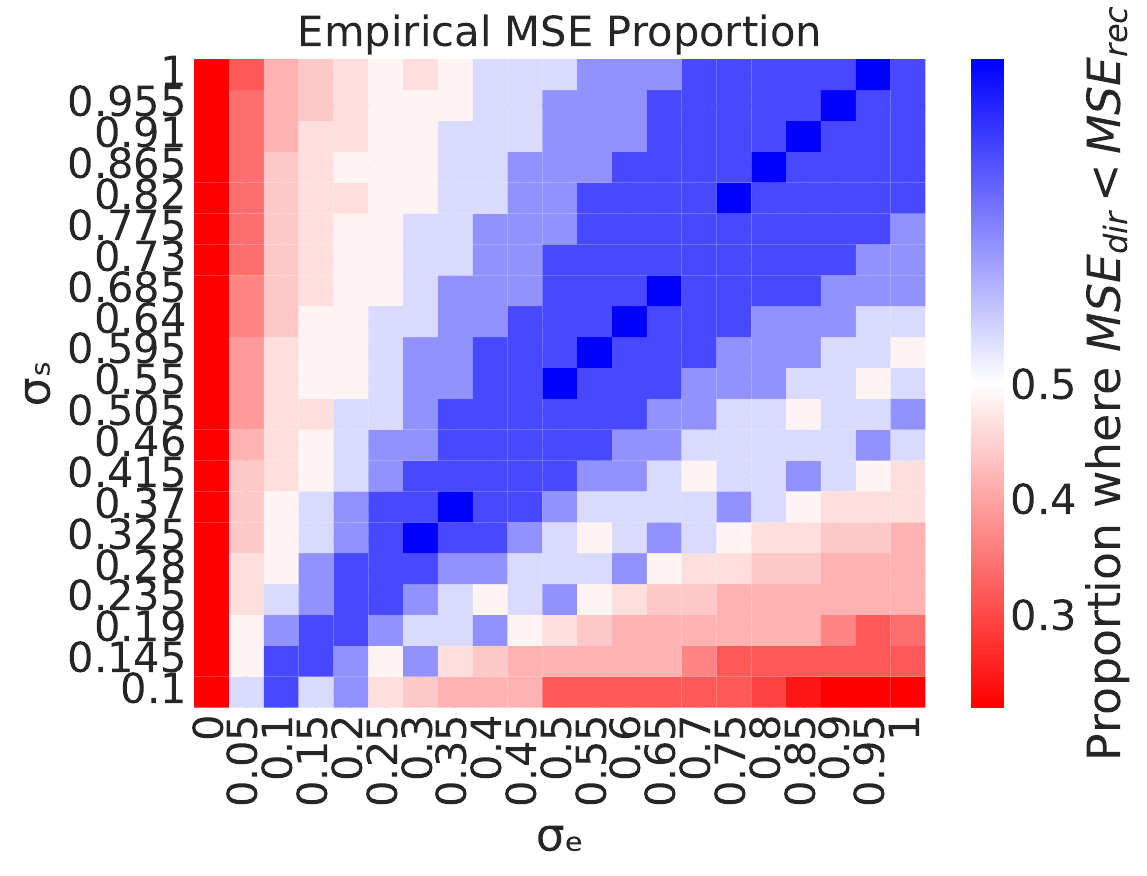}
        \caption{Empirical MSE Proportion}
        \label{fig:empirical_prop}
    \end{subfigure}
    
    \caption{Comparison of analytic and empirical proportions across $\sigma_\varepsilon$ and $\sigma_s$. 
    Left: analytic $\Delta \varepsilon$ proportion. 
    Right: empirical MSE proportion.}
    \label{fig:analytic_empirical_comparison}
\end{figure}

The empirical proportions exhibit the same qualitative behaviour predicted by the analytic expression based on the aleatoric error terms. In most regions of parameter space, the expected winner between the recursive and direct estimators follows the analytic prediction. However, for larger ratios of $\sigma_\varepsilon$ and $\sigma_s$, the correspondence weakens and noticeable deviations appear, indicating that the simplifying assumptions underlying the analytic approximation no longer hold in these limits, likely due to the OLS being biased for $\sigma_e > 0$.

\section{Appendix: Figure \ref{fig:mlp_learning_curves} Experimental Set Up
}
\label{fig_4_details}
We evaluate direct and recursive predictors on the \textbf{ETTm1} dataset using a chronological split of 60\% train and 40\% test to ensure out-of-sample generalization. 
Each model takes the most recent 50 lags as input and predicts a forecast horizon of $h{=}2$. 
Both methods share the same base MLP architecture: a single hidden layer with width~2, \texttt{tanh} activation, and a linear output layer. 
Models are trained with the Adam optimizer (learning rate $10^{-3}$) under an MSE loss. 
The \emph{recursive} model unrolls the one-step network for two steps and is trained end-to-end on the two-step objective, 
while the \emph{direct} model maps the input window directly to a two-step output vector. 
For each training sample size~$N$ in our grid and for each of 50 random seeds, both models are trained independently, 
and we record the training and test MSE. 
We report, for each~$N$, the across-seed mean and standard deviation of train/test MSE for both strategies, 
and visualize the ratios
\[
\rho_{\text{MSE}} = \frac{\text{MSE}_{\text{Rec}}}{\text{MSE}_{\text{Dir}}}, 
\qquad
\rho_{\text{Var}} = \frac{\Var_{\text{seed}}(\text{MSE}_{\text{Rec}})}{\Var_{\text{seed}}(\text{MSE}_{\text{Dir}})},
\]
where $\rho{=}1$ indicates parity, $\rho_{\text{MSE}}{<}1$ indicates lower error for the recursive model, and $\rho_{\text{Var}}{>}1$ indicates higher estimation variance.

% ======= End =======
\end{document}